\documentclass[sigconf]{acmart}
\newcommand{\good}[1]{\textcolor{green}{\textbf{#1}}}
\newcommand{\bad}[1]{\textcolor{red}{\textbf{#1}}}

\def\ie{\emph{i.e.}}
\usepackage{xspace}
\usepackage{float}
\usepackage{stfloats}
\newcommand{\name}{\textsc{Proton}\xspace}

\AtBeginDocument{%
  }
\usepackage{multirow}
\usepackage{subfigure}
\usepackage[normalem]{ulem}
\usepackage{adjustbox}

\copyrightyear{2022} 
\acmYear{2022} 
\setcopyright{rightsretained} 

\acmConference[KDD '22]{Proceedings of the 28th ACM SIGKDD Conference on Knowledge Discovery and Data Mining}{August 14--18, 2022}{Washington, DC, USA}
\acmBooktitle{Proceedings of the 28th ACM SIGKDD Conference on Knowledge Discovery and Data Mining (KDD '22), August 14--18, 2022, Washington, DC, USA}
\acmISBN{978-1-4503-9385-0/22/08}
\acmDOI{10.1145/3534678.3539305}

\usepackage{etoolbox}
\makeatletter
\patchcmd{\maketitle}{\@copyrightpermission}{
   \begin{minipage}{0.3\columnwidth}
     \href{https://creativecommons.org/licenses/by/4.0/}{\includegraphics[width=0.90\textwidth]{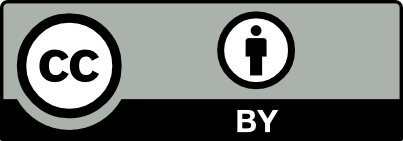}}
   \end{minipage}\hfill
   \begin{minipage}{0.7\columnwidth}
     \href{https://creativecommons.org/licenses/by/4.0/}{This work is licensed under a Creative Commons Attribution International 4.0 License.}
   \end{minipage}
  
   \vspace{5pt}
}{}{}

\makeatother

\begin{document}

\title{\name: Probing Schema Linking Information from Pre-trained Language Models for Text-to-SQL Parsing}

\author{Lihan Wang}
\authornote{Equal Contribution.}
\authornote{Lihan Wang and Bowen Qin are also with the University of Chinese Academy of Sciences. This work was conducted when Lihan Wang and Bowen Qin were interning at Alibaba.}

\affiliation{%
  \institution{Shenzhen Institute of Advanced Technology, Chinese Academy of Sciences}
  \city{Shenzhen}
  \country{China}
}

\email{lh.wang1@siat.ac.cn}

\author{Bowen Qin}
\authornotemark[1]
\authornotemark[2]
\affiliation{%
  \institution{Shenzhen Institute of Advanced Technology, Chinese Academy of Sciences}
  \city{Shenzhen}
  \country{China}
}

\email{bw.qin@siat.ac.cn}

\author{Binyuan Hui}
\authornotemark[1]
\affiliation{%
  \institution{Alibaba Group}
  \city{Beijing}
  \country{China}
}
\email{binyuan.hby@alibaba-inc.com}

\author{Bowen Li}
\affiliation{%
  \institution{Alibaba Group}
  \city{Beijing}
  \country{China}
}
\email{yanjin.lbw@alibaba-inc.com}

\author{Min Yang}
\authornote{Corresponding authors.}
\affiliation{%
  {\institution{Shenzhen Institute of Advanced Technology, Chinese Academy of Sciences}}
  \city{Shenzhen}
  \country{China}
}
\email{min.yang@siat.ac.cn}

\author{Bailin Wang}
\affiliation{%
  \institution{Massachusetts Institute of Technology}
  \city{Cambridge}
  \country{United States}
}
\email{bailinw@mit.edu}

\author{Binhua Li}
\author{Jian Sun}
\affiliation{%
  \institution{Alibaba Group}
  \city{Beijing}
  \country{China}
}

\author{Fei Huang}
\author{Luo Si}
\affiliation{%
  \institution{Alibaba Group}
  \city{Beijing}
  \country{China}
}

\author{Yongbin Li}
\authornotemark[3]
\affiliation{%
  \institution{Alibaba Group}
  \city{Beijing}
  \country{China}
}
\email{shuide.lyb@alibaba-inc.com}

\renewcommand{\shortauthors}{Lihan Wang et al.}

\begin{abstract}
The importance of building text-to-SQL parsers which can be applied to new databases has long been acknowledged, and a critical step to achieve this goal is schema linking, i.e., properly recognizing mentions of unseen columns or tables when generating SQLs. In this work, we propose a novel framework to elicit relational structures from large-scale pre-trained language models (PLMs) via a probing procedure based on Poincar\'e distance metric, and use the induced relations to augment current graph-based parsers for better schema linking. Compared with commonly-used rule-based methods for schema linking, we found that probing relations can robustly capture semantic correspondences, even when surface forms of mentions and entities differ. Moreover, our probing procedure is entirely unsupervised and requires no additional parameters. Extensive experiments show that our framework sets new state-of-the-art performance on three benchmarks. We empirically verify that our probing procedure can indeed find desired relational structures through qualitative analysis. Our code can be found at \href{https://github.com/AlibabaResearch/DAMO-ConvAI}{https://github.com/AlibabaResearch/DAMO-ConvAI}.
\end{abstract}

\ccsdesc[500]{Computing methodologies}
\ccsdesc[500]{Computing methodologies~Natural language processing}
\ccsdesc[300]{Computing methodologies~Discourse, dialogue and pragmatics}

\keywords{Semantic parsing, text-to-SQL parsing, knowledge probing}

\maketitle

\section{Introduction}
Text-to-SQL parsing aims at converting a natural language  (NL) question to its corresponding structured query language (SQL) in the context of a relational database.
Although relational databases can be efficiently accessed by skilled professionals via handcrafted SQLs, a natural language interface, whose core component relies on text-to-SQL parsing, would allow ubiquitous relational data to be accessible for a wider range of non-technical users. 
Hence, text-to-SQL parsing has attracted remarkable attention in both academic and industrial communities.

\begin{figure}
    \centering
    \includegraphics[width=0.45\textwidth]{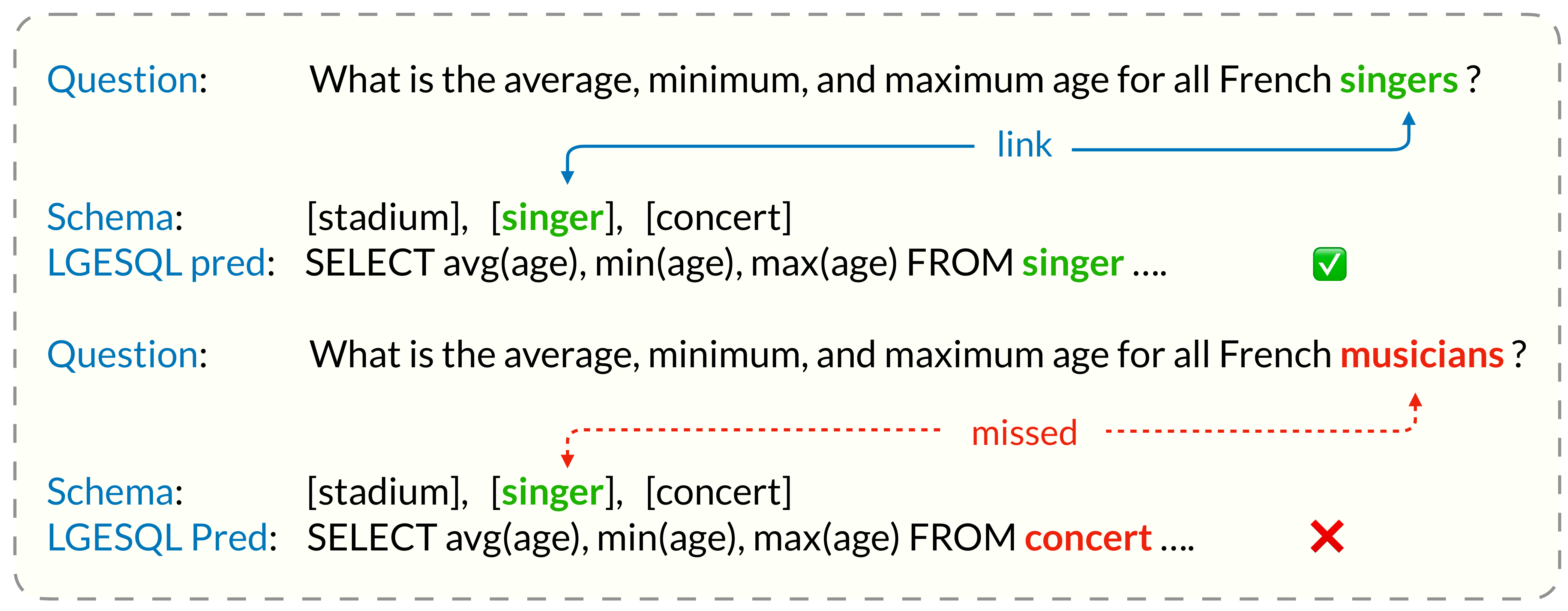}
    \caption{An illustration of the schema-linking relational structures between NL questions and database schemas. Heuristic methods such as exact string matching would not be able to capture the correspondences when surface forms of mentions appear differently.}
    \label{fig1}
\end{figure}

One challenging goal of text-to-SQL parsing is achieving \textit{domain generalization}, i.e., building parsers which can be successfully applied to new domains (or databases). 
The availability of benchmarks~\citep{wikisql,spider} has led to numerous advances  in developing parsers with domain generalization~\citep[e.g.,][]{guo2019towards,ratsql,slsql,lgesql}.
Central to achieving domain generalization is schema linking – correctly aligning entity references in the NL question to the intended schema columns or tables, even on unseen databases.
As shown in Figure~\ref{fig1}, a parser needs to link mentions \texttt{singers} and \texttt{musicians} to their corresponding column \texttt{singer}.
The importance of schema linking to domain generalization has also been verified in \cite{slsql,wang2020meta}.

Recent work suggests that standard benchmarks are limited in assessing domain generalization, and methods incorporated by current neural semantic parsers to handle schema linking cannot generalize well to more realistic settings.
Specifically, for the standard benchmark Spider~\citep{spider}, most entity mentions can be extracted by a heuristic such as string matching.~\footnote{Such artifacts might result from a biased annotation process: annotators were shown
the database schema and asked to formulate queries.} For example, 
the mention \texttt{singers} in the first question in Figure~\ref{fig1} can be trivially linked to the column \texttt{singer} based on their surface forms. Many  parsers~\citep{guo2019towards,ratsql} exploit such artifacts (or shortcuts), but their good performance on Spider does not transfer to real-life settings where mentions and columns/tables are very likely to share different surface forms. For example, when the mention \texttt{singer} is replaced with its synonym \texttt{musicians}, the state-of-the-art parser LGESQL~\citep{lgesql} fails to handle the schema-linking relations.

In this work, we propose a novel approach to elicit relational structures for schema linking from large-scale pre-trained language models (PLMs) through a probing procedure.
In addition to simply encoding NL question and schema in  \textit{continuous} vector space using PLMs, as most previous semantic parsers do, we propose to distill \textit{discrete} relational structures from PLMs. Such relational structures are extracted in an unsupervised manner, and they can be directly exploited for schema linking when generating structured programs of SQLs. We capitalize on the intuition that although relational information is already contained in the continuous representations of PLMs, neural parsers lack an optimal mechanism to benefit from such information. The algorithmic inductive biases introduced by our probing procedure would allow the underlying relational structures to be explicitly and easily exploited by neural parsers. Previous work has shown that PLMs such as BERT \cite{devlin2018bert}, RoBERTa \cite{liu2019roberta}, ELECTRA \cite{clark2020electra} store linguistic knowledge \cite{hewitt-liang-2019-designing}, world knowledge \cite{raffel2019exploring} and relational knowledge \cite{petroni2019language}. To our best knowledge, we are the first work to adapt probing methods to exploit relational information from PLMs for the complex structured prediction task of text-to-SQL parsing.

The relational structures extracted from PLMs hold several appealing properties which make them suitable for domain generalization of text-to-SQL parsing. First, they are domain-invariant, and this is inherited from that PLMs are usually obtained via self-supervised training on various domains of textual data. Second, they can better capture semantic correspondences than current heuristics such as n-gram matching~\citep{spiderSYN,ratsql}.
Third, they are relatively more robust to the cross-database setting. As they are elicited in an unsupervised probing procedure and not induced during in-domain training, they will not suffer from overfitting to observed training databases.

In this work, we propose a novel framework, called \textbf{\name}\footnote{\underline{\textbf{PRO}}bing Schema Linking Informa\underline{\textbf{T}}i\underline{\textbf{O}}n from Pre-trai\underline{\textbf{N}}ed Language Models}, which first probes the underlying relational schema-linking structures between a NL query and its database schema from a pre-trained language model, and then effectively injects it into the downstream text-to-SQL parsing models.
To better model the heterogeneous relational structures, we introduce a probing procedure based on Poincar\'e distance metric instead of the traditional Euclidean distance metric, inspired by \citet{chen2021probing}.
Our probing procedure is entirely unsupervised and does not require additional parameters.
We empirically show the effectiveness of \textbf{\name} on several text-to-SQL benchmarks, \ie, Spider~\cite{spider}, SYN~\cite{spiderSYN} and DK~\cite{spiderDK}, and through qualitative analysis, we verify that our probing procedure can indeed find desired relational structures.

The contributions of this work can be summarized as follows:
\begin{itemize}
    \item To boost domain generalization for text-to-SQL parsing, we propose a novel framework that utilizes relational schema-linking structures that are extracted from a PLM via an unsupervised probing process. 
    \item  To better capture the heterogeneous relational structures between NL queries and database schema, we introduce a Poincar\'e distance metric that can better measure semantic relevance than the typical Euclidean distance metric.
    \item Extensive experiments on three text-to-SQL benchmarks show that our probing method can lead to significantly better results when compared with current state-of-the-art parsers.
\end{itemize}

\begin{figure*}
    \centering
    \setlength{\abovecaptionskip}{5pt} 
    \includegraphics[width=0.7\textwidth]{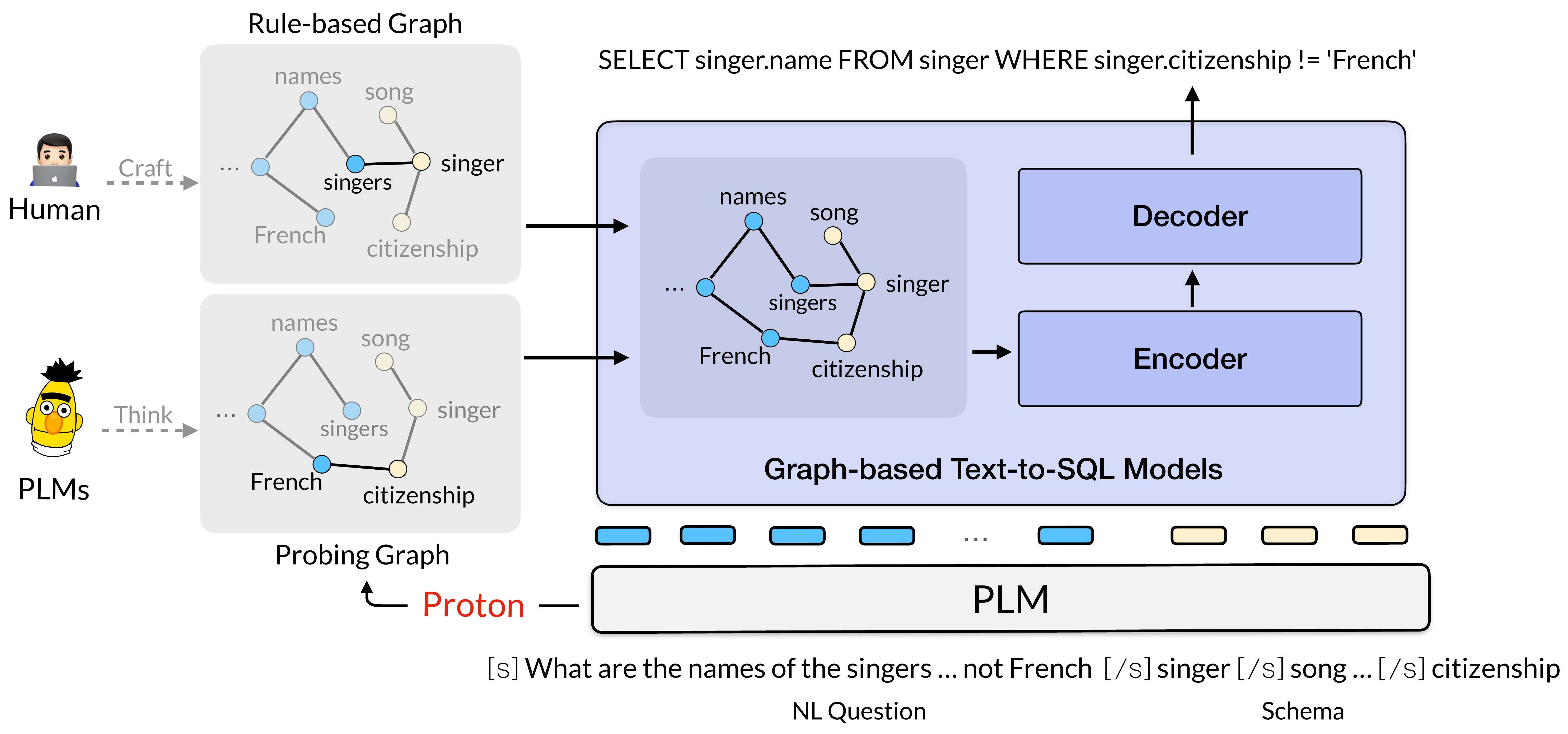}
    \caption{The overview of our proposed framework for text-to-SQL parsing. To obtain relation graphs among NL questions and database schema, we first use our proposed method \name to probe relational structures from PLMs. The induced relations, along with the commonly-used heuristic relations extracted with handcrafted rules, are then utilized by a graph-based text-to-SQL parser to boost its schema linking for domain generalization. 
    }
    \label{pipeline}
\end{figure*}

Though we only focus on text-to-SQL parsing in this work, we believe that the general methodology of probing underlying discrete relational structures from PLMs can be extended to related tasks that require complex reasoning over structured knowledge, such as knowledge-based question answering~\citep{gu2021beyond} and dialog~\citep{dai2020learning,dai2021preview,lin2022duplex}, table-based fact checking~\citep{chen2019tabfact} and structured data record to text generation~\citep{nan2020dart}. 

\section{Graph-based Text-to-SQL Models}
\label{sec:graph}

\subsection{Notation Definition}
Given a natural language question $Q$  and the corresponding database schema $\mathcal{S}=\langle\mathcal{T}, \mathcal{C}\rangle$, the target is to generate a SQL query $Y$.
More specifically, the question $Q = \left\{q_{1}, q_{2}, \cdots, q_{|Q|} \right\}$ is a sequence of tokens, and the schema consists of tables $T=\left\{t_{1}, t_{2}, \cdots, t_{|\mathcal{T}|} \right\}$ and columns $C=\left\{c_{1}, c_{2}, \cdots, c_{|\mathcal{C}|} \right\}$.
Each table $t_i$ contains multiple words $(t_{i,1}, t_{i,2}, \cdots, t_{i,|t_i|})$ and each column name $c_j^{t_i}$ in table $t_i$ contains words $(c_{j,1}^{t_{i}}, c_{j,2}^{t_{i}}, \cdots, c_{j,|c_j^{t_i}|}^{t_{i}})$. Formally, we denote the input  as $X$, where $X = \langle\mathcal{Q}, \mathcal{S}\rangle$ and the desired SQL query as $Y$ which is represented as an abstract syntax tree (AST) \cite{YinN17} in the context-free grammar of SQL.
We employ the de facto encoder-decoder framework, where the encoder jointly maps NL questions and schema items into embeddings $\mathbf{X}$ and the decoder generates the AST of the target query $Y$ in the depth-first-search order. In this paper, we adopt two representative graph-based models RAT-SQL \cite{ratsql} and LGESQL \cite{lgesql} as our base models given their SOTA performance. 

\subsection{Encoder}
Formally, the graph-based models RAT-SQL and LGESQL consider the NL question and the database schema as a single direct graph $\mathcal{G} = \langle \mathcal{V}, \mathcal{E} \rangle $, where $\mathcal{V} = Q \cup \mathcal{T} \cup \mathcal{C}$ denotes the node set containing the input NL question tokens as well as schema items and $\mathcal{E}$ is the edge set depicting  \emph{pre-existing} relations between NL question tokens and schema items. 
Given the inputs $X=\left\{\boldsymbol{x}_{\boldsymbol{i}}\right\}_{i=1}^{n}$ and input graph $\mathcal{G}$,  a relational-aware transformer (RAT) \cite{ratsql} is leveraged as the encoder.
The relation-aware transformer is based on the classic transformer but represents relative position information in a self-attention layer, which transforms each $\boldsymbol{x}_{\boldsymbol{i}}$ into $\boldsymbol{y}_{\boldsymbol{i}} \in \mathbb{R}^{d_{\boldsymbol{x}}}$ as follows:
\begin{equation}
e_{i j}^{(h)} =\frac{\boldsymbol{x}_{i} W_{Q}^{(h)}\left(\boldsymbol{x}_{j} W_{K}^{(h)}+r_{i j}\right)^{\top}}{\sqrt{d_{z} / H}}
\label{rij1}
\end{equation}
\vspace{-0.1cm}
\begin{equation}
 \alpha_{i j}^{(h)}=\operatorname{softmax}_{j}\left\{e_{i j}^{(h)}\right\} 
\end{equation}
\vspace{-0.2cm}
\begin{equation}
    \boldsymbol{z}_{i}^{(h)} =\sum_{j=1}^{n} \alpha_{i j}^{(h)}\left(\boldsymbol{x}_{j} W_{V}^{(h)}+r_{i j}\right) \label{rij2}
\end{equation}
\vspace{-0.2cm}
\begin{equation}
    \boldsymbol{z}_{i}=\operatorname{Concat}\left(\boldsymbol{z}_{i}^{(1)}, \cdots, \boldsymbol{z}_{i}^{(H)}\right)
\end{equation}
\vspace{-0.2cm}
\begin{equation}
    \tilde{\boldsymbol{y}}_{\boldsymbol{i}} =\operatorname{LayerNorm}\left(\boldsymbol{x}_{\boldsymbol{i}}+\boldsymbol{z}_{i}\right)
\end{equation}
\vspace{-0.3cm}
\begin{equation}
    \boldsymbol{y}_{\boldsymbol{i}}  =\operatorname{LayerNorm}\left(\tilde{\boldsymbol{y}}_{\boldsymbol{i}}+\operatorname{FC}\left(\operatorname{ReLU}\left(\operatorname{FC}\left(\tilde{\boldsymbol{y}}_{\boldsymbol{i}}\right)\right)\right)\right.
\end{equation}
where FC is a fully-connected layer and LayerNorm is the layer normalization operation \cite{ba2016layer}.
$W_{Q}^{(h)}, W_{K}^{(h)}, W_{V}^{(h)} \in \mathbb{R}^{d_{x} \times\left(d_{x} / H\right)}$ are learnable parameters where $d_{x}$ denotes the dimension of hidden representation.
$d_{x}$ and $d_{z}$ represents the dimension of $x$ and $z$.
$H$ is the number of heads and we have $1 \leq h \leq H$. 
Here, the term $r_{ij}$ encodes the known relationship between the two elements $x_i$ and $x_j$ in the input.
The RAT framework represents all the pre-existing features for each edge $(i,j)$ as
$r_{i j}$ in which each element is either a learnable embedding for each corresponding edge or a zero vector if the relation does not hold for the edge.
The reader can refer to \cite{ratsql} for the implementation details of RAT.

LGESQL applies a line-graph enhanced relational graph attention network (RGAT) as its encoder.
Different from RAT, RGAT is based on graph attention network and represents relative position information in a self-attention layer.
Compared with normal RGAT, line-graph enhanced RGAT employs an additional edge-centric line graph constructed from the original node-centric graph. During the iteration process of node embeddings, each node in either graph integrates information from its neighborhood and incorporates edge features from the dual graph to update its representation. 
Due to the limited space, we omit the formal definition of RGAT.
The reader can refer to \cite{lgesql} for the implementation details of LGESQL.

\subsection{Decoder}
Both RAT-SQL and LGESQL apply grammar-based syntactic neural decoder \cite{YinN17} to generate the abstract syntax tree (AST) of the target query $Y$ in a depth-first-search order. 
The output at each decoding time step is either 1) an  $\mathtt{APPLYRULE}$ action that expands the current non-terminal node in the partially generated AST, or 2) $\mathtt{SELECTTABLE}$ or $\mathtt{SELECTCOLUMN}$ action that chooses one schema item from the output memory of encoder. 
The readers can refer to \cite{ratsql} for more details.

\paragraph{\textbf{Discussion of Schema Linking}}
As mentioned above, $r_{i,j}$ in Eq.\ref{rij1} and Eq.\ref{rij2} represents the schema linking items in inputs.
The graphs adopted in RAT-SQL and LGESQL are constructed by schema linking with lexical matching. 
For instance, the word ``\textit{cylinders}'' in a NL question will be linked to the \texttt{cylinders} column in a table \texttt{cars\_data}.
In this way, a relation-aware input graph can be constructed, which is represented as an adjacency matrix.
However, the rule-based string matching is inapplicable in more challenging scenarios where the NL questions contain implicit mentions such as synonym substitution and entity abbreviation. 
The missing linkage may hinder the encoder's ability to capture salient relations.
In this paper, we propose to probe schema linking information from large-scale PLMs that are claimed to contain rich \emph{semantic} relational knowledge implicitly.
It is noteworthy that our probing technique is model-agnostic and potentially applicable for any text-to-SQL parsing models. In the next section, we will introduce the details of our probing technique.

\section{Probing Schema Linking}
\begin{figure}
    \centering
    \includegraphics[width=0.3\textwidth]{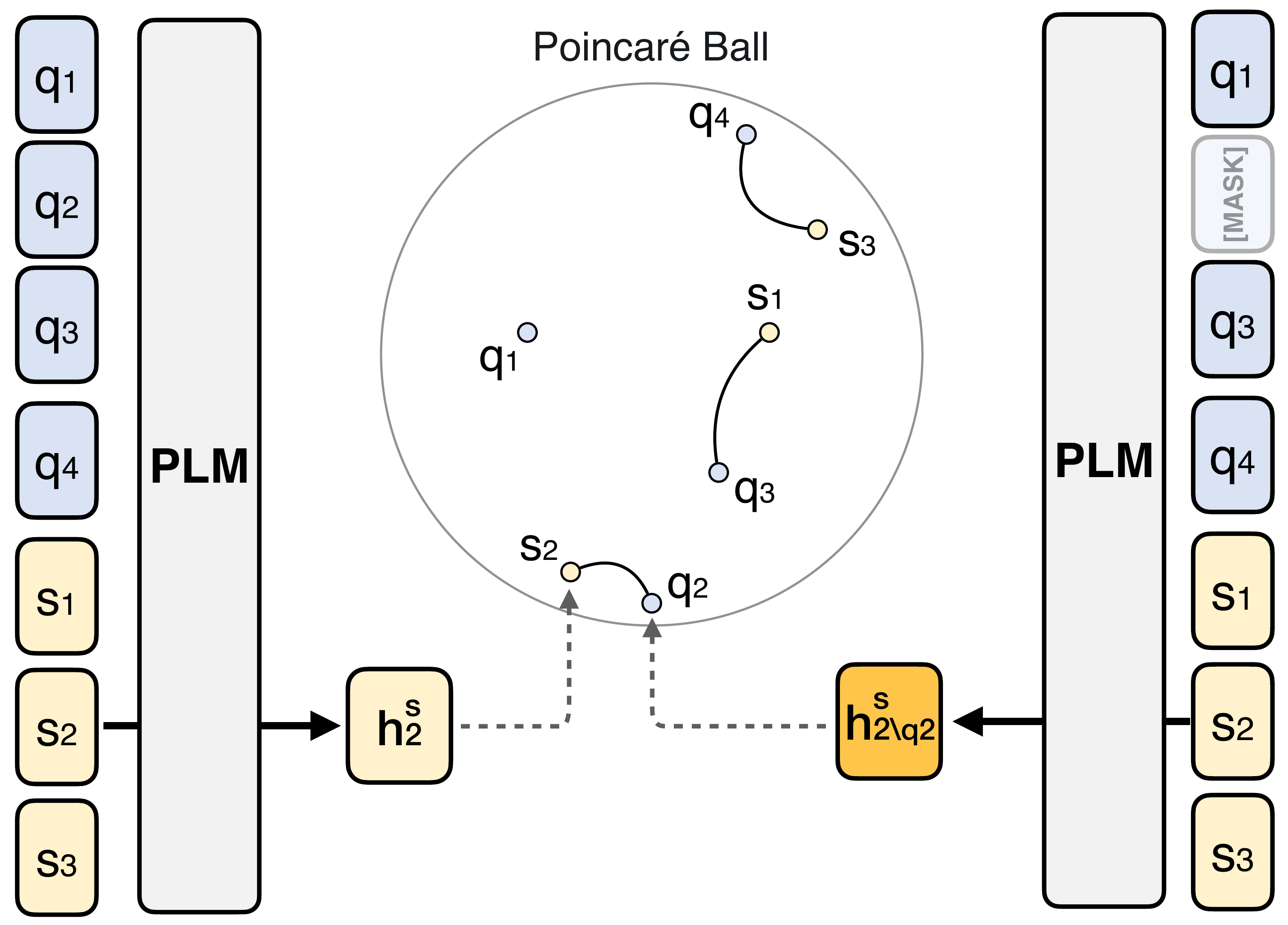}
    \caption{
    Probing process for \name.
    $\mathbf{h}^s_2$ represents the embedding of the schema item $s_2$, while $\mathbf{h}^s_{2\backslash q_2}$ represents the embedding when the question token $q_2$ is masked out.
    The relation between $s_2$ and $q_2$ as well as other candidate pairs are identified in the Poincar\'e ball.}
    \label{prob}
\end{figure}

In this section, we introduce a  parameter-free probing technique, as illustrated in Figure \ref{prob}, to probe schema linking information between the NL query and the database schema from PLMs. Concretely, we propose a masking technique to measure the correlation between NL question tokens and schema items (i.e., columns and tables) in the masked language modeling (MLM) process. 

\subsection{Probing Stage}

Given a database schema $\mathcal{S}=\langle\mathcal{T}, \mathcal{C}\rangle$, where the table and column sequences are $\mathcal{T}=\left(t_{1}, t_{2}, \cdots, t_{|\mathcal{T}|} \right)$ and $\mathcal{C}=\left(c_{1}, c_{2}, \cdots, c_{|\mathcal{C}|} \right)$ respectively.
We first concatenate $T$ and $C$ into a single sequence $\mathcal{S}=\left(\mathcal{T}, \mathcal{C}\right)=\left(s_1, s_2, ..., s_{|\mathcal{T}| + |\mathcal{C}|}\right)$.
Together with the NL question sequence $\mathcal{Q}=\left(q_1, q_2, ..., q_{|Q|}\right)$, the input $\mathcal{I}$ is formed by a sequential concatenation of $\mathcal{Q}$ and $\mathcal{S}$ as:
\begin{equation*}
\resizebox{1\hsize}{!}{$
\mathcal{I}= (\langle\mathsf{s}\rangle;q_1;\dots;q_{|Q|};\langle\backslash\mathsf{s}\rangle; s_1;\langle\backslash\mathsf{s}\rangle;\dots;\langle\backslash\mathsf{s}\rangle; s_{|\mathcal{T}| + |\mathcal{C}|}),
 $}
\end{equation*}
where $\langle\mathsf{s}\rangle$, $\langle\backslash\mathsf{s}\rangle$ are special tokens to delimit the input tokens. 
The MLM maps the input $\mathcal{I}$ into the deep contextualized representations. 
We denote $(\mathbf{h}^{q}_1, \ldots, \mathbf{h}^{q}_{|Q|})$ and $(\mathbf{h}^{s}_1, \ldots, \mathbf{h}^{s}_{|\mathcal{T}| + |\mathcal{C}|})$ as  question token representations and schema item representations, respectively.

The goal of our probing technique is to derive a function $f(\cdot,\cdot)$ that captures the correlation between an arbitrary pair of a question token and a schema item.
To this end, we employ a two-step MLM process.
It is inspired by the observation that a word is considered as essential for document classification if removing the word from a document leads to a considerable accuracy decrease.

As shown in Figure \ref{prob}, we first feed the input $\mathcal{I}$ into the PLM. We use $\mathbf{h}^{s}_{j}$ to denote the contextualized representation of the $j$-th schema item $s_j$, where $1 \le j \le |\mathcal{T}| + |\mathcal{C}|$.
Then, we replace the question token $q_i$ with a special mask token $\texttt{[MASK]}$ and feed the corrupted input $\mathcal{I}\backslash\left\{q_i\right\}$ into the PLM again.
Accordingly, we use $\mathbf{h}^{s}_{j \backslash q_i}$ to denote the new representation of the $j$-th schema item when $q_i$ is masked out.\footnote{
\citet{wu2020perturbed} proposed a similar two step perturbed masking.
The difference is that they first masked out one input token $\{s_i\}$ and then masked out a token pair $\{s_i, s_j\}$.
Then the correlation was obtained by comparing $\mathbf{h}_{i \backslash \{s_i\}}$ and $\mathbf{h}_{i \backslash \{s_i, s_j\}}$.
This method heavily relies on the MLM prediction ability of the PLMs (e.g., BERT).
We tried it in our preliminary experiments, but it performed poorly.
We conjecture that our input data (i.e., questions and schemas) differs from the PLM's pre-training data, and the PLM struggles to make reasonable schema predictions without further finetuning.
}

Formally, we measure the distance between $\mathbf{h}^{s}_{j}$ and $\mathbf{h}^{s}_{j \backslash q_i}$ to induce the correlation between the schema item $s_j$ and the question token $q_i$ as follows:
\begin{equation}
	f(q_i, s_j) = d(\mathbf{h}^{s}_{j \backslash q_i}, \mathbf{h}^{s}_{j})
\end{equation}
where $d(\cdot,\cdot)$ is the distance metric to measure the difference between two vectors. Generally, we can use Euclidean distance metric to implement  $d(\cdot,\cdot)$:
\begin{equation}
\label{eq:euclidean}
	d_{\rm Euc}(q_i, s_j) = || \mathbf{h}^{s}_{j \backslash q_i} - \mathbf{h}^{s}_{j}||_2
\end{equation}
where $d_{\rm Euc}(\cdot,\cdot)$ denotes a distance function in Euclidean space.

\subsection{Poincar\'e Probe}
Euclidean space has intrinsic difficulties in modeling complex data \cite{bronstein2017geometric}. 
To better model the heterogeneous relational structures between the NL query and the database schema, we devise a Poincar\'e probe, which probes schema linking information from PLMs in the hyperbolic space that is expected to better capture linguistic hierarchies encoded in contextualized representations \citep{nickel2017poincare, Tifrea19PoincareGlove}. As revealed in \cite{chami2019hyperbolic}, the hyperbolic space enables vector comparison with much smaller distortion compared with the Euclidean space. In addition, recent work \cite{krioukov2010hyperbolic,nickel2017poincare,chami2019hyperbolic} demonstrates that the hyperbolic space may reflect some properties of graph naturally.

\paragraph{\textbf{The Poincar\'e Ball}}
In this paper, we employ the standard Poincar\'e ball, which is a special model of hyperbolic spaces, to capture the difference between $\mathbf{h}^{s}_{j \backslash q_i}$ and  $\mathbf{h}^{s}_{j}$.
Before introducing the Poincar\'e Probe, we first review basic concepts of the standard Poincar\'e ball following \citet{ganea2018hyperbolic}. Formally, for a point $\mathbf{x}$ in the hyperbolic space, the standard Poincar\'e ball model is defined as $\left(\mathbb{D}^{n},g_{\mathbf{x}}^{\mathbb{D}}\right)$, where $\mathbb{D}^{n}=\left\{\mathbf{x} \in \mathbb{R}^{n} \mid \|\mathbf{x}\|^{2}<1\right\}$ is a Riemannian manifold and $g_{\mathbf{x}}^{\mathbb{D}}=\left(\lambda_\mathbf{x}\right)^{2} \mathbf{I}_{n}$ is the metric tensor. We formulate $\lambda_{\mathbf{x}}=2 /\left(1-\|\mathbf{x}\|^{2}\right)$ as the conformal factor. Here, $n$ denotes the dimension size. 

\paragraph{\textbf{Feature Projection}}
To compare the feature vectors learned by PLMs in the hyperbolic space, we first use the exponential mapping function $g_\mathbf{x}(\cdot)$ to project the embeddings into the hyperbolic space. 
Suppose $\mathbf{h}\in \mathbb{T}^n_{\mathbf{x}}$ is the input vector in the tangent space with respect to the point $\mathbf{x}$ in the hyperbolic space. Here, $\mathbf{h}$ will be instantiated as $\mathbf{h}^{s}_{j \backslash q_i}$ and $\mathbf{h}^{s}_{j}$. The mapping function $g_\mathbf{x}(\cdot)$: $\mathbb{T}_{\mathbf{x}}^{n} \rightarrow \mathbb{D}^{n}$ can be computed by:
\begin{equation}
  g_{\mathbf{x}}\left(\mathbf{h}\right)= \mathbf{x} \oplus \left(\tanh \left( \frac{\lambda_{x}\left\|\mathbf{h}\right\|}{2}\right) \frac{\mathbf{h}}{\left\|\mathbf{h}\right\|}\right) 
\end{equation}
where the operation $\oplus$ is the M\"obius addition. For any $\mathbf{a}, \mathbf{b} \in \mathbb{D}^{n}$, it is calculated as:
\begin{equation}
\label{eq:mobius}
 \mathbf{a} \oplus \mathbf{b} = \frac{
 \left(1+2\left\langle \mathbf{a}, \mathbf{b}\right\rangle+\left\|\mathbf{b}\right\|^{2}\right) \mathbf{a}+\left(1-\left\|\mathbf{a}\right\|^{2}\right) \mathbf{b} }{ 1+2 \left\langle \mathbf{a}, \mathbf{b}\right\rangle+\left\|\mathbf{a}\right\|^{2}\left\|\mathbf{b}\right\|^{2}} 
\end{equation}

In this work, we assume that $\mathbf{h}$ lies in the tangent space at the point $\mathbf{x}=\mathbf{0}$.
Then, the hyperbolic representation $\tilde{\mathbf{h}}$ of $\mathbf{h}$ can be obtained by:
\begin{equation}
\label{eq:exp proj}
\tilde{\mathbf{h}} =  g _{\mathbf{0}}\left(\mathbf{h}\right)=\tanh \left( \left\|\mathbf{h}\right\|\right) \frac{\mathbf{h}}{  \left\|\mathbf{h}\right\|}.
\end{equation}

\paragraph{\textbf{The Poincar\'e Distance}}
After obtaining the feature representations in the hyperbolic space via the feature mapping function $g_{\mathbf{x}}(\cdot)$, we can measure the correlation between the schema item $s_j$ and the question token $q_i$ by calculating the Poincar\'e distance between  $\tilde{\mathbf{h}}^s_j$ and $\tilde{\mathbf{h}}^s_{j\backslash {q_i}}$ in the hyperbolic space as follows:
\vspace{-0.15cm}
\begin{equation}
	d_{\rm Poin}(q_i, s_j) = 2 \tanh ^{-1} ( \| - \tilde{\mathbf{h}}^s_{j\backslash {q_i}}  \oplus \tilde{\mathbf{h}}^s_j \|)
\end{equation}
where $\oplus$ is the M\"obius addition defined in Eq. (\ref{eq:mobius}). We can replace the Euclidean distance $d_{\rm Euc}(\cdot, \cdot)$ defined in Eq. (\ref{eq:euclidean}) with the Poincar\'e distance $d_{\rm Poin}(\cdot, \cdot)$ to implement the function $f(\cdot, \cdot)$ for relational knowledge probing.

\subsection{Schema Linking for Graph Construction}
By repeating the two-stage MLM process on
each pair of tokens $q_i$, $s_j$ and calculating 
$f(q_i, s_j)$, we obtain a relation matrix $\mathbf{X} = \{x_{i,j}\}_{i=1, j=1}^{|Q|,|S|}$, where $x_{i, j}$ denotes the relation between question-schema pair $(q_i, s_j)$. 
We utilize the min-max normalization to reduce the impact of the range of correlation scores: 
\begin{equation}
\tilde{x}_{i,j} = \frac{x_{i,j}-\min \left(\mathbf{X}\right)}{\max \left(\mathbf{X}\right)-\min \left(\mathbf{X}\right)}
\end{equation}

Now, we can derive a strategy to construct the unweighted direct graph $G$ used in RAT-SQL and LGESQL. Specifically, we convert the relation matrix $\mathbf{X}$ into an adjacency matrix $\mathbf{A}$ that represents the structure of graph $\mathcal{G}$. We compute the adjacency matrix $\mathbf{A}$ by:
\begin {equation}
\label{adjacency}
\mathbf{A}_{i j}=\left\{\begin{array}{cl}
\text {0} & \text { if } \tilde{x}_{i, j} < \tau \\
\text { 1 } & \text { if } \tilde{x}_{i, j} > \tau 
\end{array}\right. ,
\end{equation}
where $\tau$ is a pre-defined threshold.
The learned adjacency matrix $\mathbf{A}$ via probing the relational knowledge from PLMs can facilitate the semantic linking of text-to-SQL parsing. 

\section{Experimental Setup}
\subsection{Datasets}
We conduct extensive experiments on three benchmark  datasets. (1) \textbf{Spider} \cite{spider} is a large-scale cross-domain zero-shot text-to-SQL benchmark.
We follow the common practice to report the exact match accuracy on the development set, as the test set is not publicly available. (2) \textbf{DK} \cite{spiderDK} is a human-curated dataset based on  Spider, a challenging variant of the Spider development set, with focus on evaluating the model understanding of domain knowledge. 
(3) \textbf{SYN} \cite{spiderSYN} is another challenging variant of Spider.
SYN is constructed by manually modifying NL questions in Spider using synonym substitution, which aims to simulate the scenario where users do not know the exact schema words in the utterances.

\subsection{Baselines}
We choose RAT-SQL and LGESQL as our base parsers,
where RAT-SQL is a sequence-to-sequence model enhanced by a relational-aware transformer and LGESQL is a graph attention network based sequence-to-sequence model with the relational GAT and the line graph.
Both models use schema linking to build the input graph.
Meanwhile, for a comprehensive comparison, we also compare our model with several recent state-of-the-art models, including GNN \citep{bogin-representing}, IRNet \citep{guo2019towards}, EditSQL \citep{editsql}, RYANSQL \citep{Choi2020RYANSQLRA}, TranX \citep{Yin2020TaBERTPF}. 
For RAT-SQL, we choose RAT-SQL + Grappa as our baseline, which has the best performance of RAT-SQL.
We adopt the same hyper-parameter settings as in \citet{Yu2020GraPPaGP}.

\subsection{Implementation Details}
For RAT-SQL, in the encoder, the hidden size of bidirectional LSTMs per direction is set to 128. 
The number of relation-aware self-attention layers stacked on top of the bidirectional LSTMs is 8 and the dimension of each attention layer is 256.
The position-wise feed-forward network has inner layer dimension 1024. In the decoder, the dimension of rule embedding, node type embedding and hidden state are set to 128, 64 and 512 respectively. 
RAT-SQL applies Adam optimizer \cite{kingma2014adam} with default hyperparameters.
The batch size is 8 and the number of training steps is 40,000.

For LGESQL, in the encoder, the GNN hidden size is set to 512 for PLMs. The number of GNN layers is 8. In the decoder, the dimension of hidden state, action embedding and node type embedding are set to 512, 128 and 128 respectively. 
The recurrent dropout rate is 0.2 for decoder LSTM. The number of heads in multi-head attention is 8 and the dropout rate of features is set to 0.2 in both the encoder and decoder.
LGESQL uses AdamW optimizer \cite{loshchilov2017decoupled} with linear warmup scheduler and the warmup ratio of total training steps is 0.1.
The maximum gradient norm is set to 5.
The batch size is 20 and the number of training epochs is 200.

\begin{table}[t]  
    \centering
     \setlength{\abovecaptionskip}{5pt} 
    \caption{Exact match accuracy (\%) on DK benchmark. }
    \small
    \begin{tabular}{lc}  
    \toprule
    \textbf{Model}& \textbf{Acc.} \\ 
    \midrule
    GNN + BERT \citep{bogin-representing} & 26.0 \\ 
    IRNet + BERT \citep{guo2019towards} & 33.1  \\ 
    RAT-SQL  \citep{ratsql} & 35.8\\
    RAT-SQL + BERT \citep{ratsql} & 40.9 \\
    RAT-SQL + GAP \citep{Shi2020LearningCR} & 44.1\\
    
    \midrule
    
    RAT-SQL + Grappa  & 38.5  \\
    \quad w/ Euclidean \name & {45.0} ($\uparrow$ 6.5)\\
    \quad w/ Hyperbolic \name& \textbf{46.4} ($\uparrow$ \textbf{7.9})\\
    \midrule
    
    LGESQL + RoBERTa-large & 45.9   \\
    \quad w/ Euclidean \name& {46.2} ($\uparrow$ 0.3)\\
    \quad w/ Hyperbolic \name& \textbf{46.7} ($\uparrow$ 0.8)\\
    \midrule
    LGESQL + ELECTRA-large & 48.4   \\
    \quad w/ Euclidean \name& {49.3} ($\uparrow$ 0.9)\\
    \quad w/ Hyperbolic \name& \textbf{51.0} ($\uparrow$ 2.6)\\
    
    \bottomrule
    \end{tabular}  
    \label{tab:dk}
    \vspace{-0.3cm}
\end{table}

\begin{table}[t]  
    \centering
    \setlength{\abovecaptionskip}{5pt}  
    \caption{Exact match accuracy (\%) on SYN benchmark. }
    \small
    \begin{tabular}{lc}  
    \toprule
    \textbf{Model}& \textbf{Acc.} \\ 
    \midrule
    GNN \citep{bogin-representing} & 23.6 \\ 
    IRNet  \citep{guo2019towards} & 28.4  \\ 
    RAT-SQL  \citep{ratsql} & 33.6\\
    RAT-SQL + BERT \citep{ratsql} & 48.2\\
    \midrule 
    
    RAT-SQL + Grappa & 49.1\\

    \quad w/ Euclidean \name& {61.4} ($\uparrow$ 12.3)\\
    \quad w/ Hyperbolic \name& \textbf{62.6} ($\uparrow$ \textbf{13.5})\\
    \midrule
    LGESQL + RoBERTa-large & 54.1    \\

    \quad w/ Euclidean \name& {57.7} ($\uparrow$ 3.6)\\
    \quad w/ Hyperbolic \name& \textbf{58.6} ($\uparrow$ 4.5)\\
    \midrule
    LGESQL + ELECTRA-large & 64.6 \\
    \quad w/ Euclidean \name& {65.4} ($\uparrow$ 0.8)\\
    \quad w/ Hyperbolic \name& \textbf{65.6} ($\uparrow$ 1.0)\\
    
    \bottomrule
    \end{tabular}  
    \label{tab:syn}
    \vspace{-0.3cm}
\end{table}

\section{Experimental Results}
\subsection{Main Results}
Tables \ref{tab:dk}-\ref{tab:spider} show the results of DK, SYN and Spider datasets, respectively. From the results, we have the following observations. First, we can observe that the LGESQL with our probing methods (\ie, Euclidean \name and Hyperbolic \name) yields substantially better results than the compared baseline methods on all the datasets. In particular,  LGESQL+ELECTRA-large with Hyperbolic \name achieves the exact match accuracy of 51.0\% on the DK benchmark, which obtains 2.6\% improvement over LGESQL+ELECTRA-large. 
The effectiveness of \name on the DK benchmark demonstrates that the semantic and relational knowledge elicited from PLMs can help text-to-SQL parsers understand the domain knowledge and adapt the knowledge to a new domain. Similar trends can be observed on the SYN benchmark and the Spider dataset. For instance, LGESQL+ELECTRA-large with Hyperbolic \name achieves the strong performance (76.3\%) for text-to-SQL parsing on the Spider dataset. The exact match accuracy increases by 1.0\% over the best baseline LGESQL+ELECTRA-large.  As we know, it is difficult to boost 1\% of exact match accuracy on the Spider dataset.  This also verifies the effectiveness of our \name model.
Secondly, we find that Hyperbolic \name performs better than Euclidean \name with notable improvements across all datasets. This is because the Poincar\'e distance metric can better measure semantic relevance between NL queries and database schema than the typical Euclidean distance metric.
Third, Hyperbolic \name can largely improve the performance of RAT-SQL+Grappa, up to 7.9\% on the DK benchmark and 13.5\% on SYN benchmark. The reason may be that RAT-SQL+Grappa is not well designed for schema linking, while our Hyperbolic \name can effectively capture the relational information between the NL query and the database schema by probing relational knowledge from PLMs. 

\begin{table}[t]  
    \centering
    \setlength{\abovecaptionskip}{5pt} 
    \caption{Exact match accuracy (\%) on Spider dev set. The $*$ means re-implemented results.}
    \small
    \begin{tabular}{lc}  
    \toprule
    \textbf{Model}& \textbf{Acc.} \\ 
    \midrule
    EditSQL + BERT \citep{editsql} & 57.6 \\ 
    IRNet + BERT \citep{guo2019towards} & 61.9  \\ 
    RYANSQL + BERT \citep{Choi2020RYANSQLRA} & 70.6  \\
    TranX + TaBERT \citep{Yin2020TaBERTPF} & 65.2  \\  
    RAT-SQL + BERT \citep{ratsql} & 69.7 \\
    \midrule
    RAT-SQL + Grappa$^{*}$  & 71.2 \\
    \quad w/ Euclidean \name& 72.6 ($\uparrow$ 1.4)\\
    \quad w/ Hyperbolic \name& \textbf{73.1} ($\uparrow$ \textbf{1.9})\\
    
    \midrule
    LGESQL + RoBERTa-large$^{*}$  & 71.7   \\
    \quad w/ Euclidean \name& {72.9} ($\uparrow$ 1.2)\\
    \quad w/ Hyperbolic \name& \textbf{73.3} ($\uparrow$ 1.6)\\
    \midrule
    LGESQL + ELECTRA-large & 75.3  \\

    \quad w/ Euclidean \name& {76.0} ($\uparrow$ 0.7)\\
    \quad w/ Hyperbolic \name& \textbf{76.3} ($\uparrow$ 1.0)\\
    
    \bottomrule
    \end{tabular}  
    \label{tab:spider}
\end{table}

\begin{figure*}
    \centering
    \setlength{\abovecaptionskip}{5pt} 
    \includegraphics[width=0.8\textwidth]{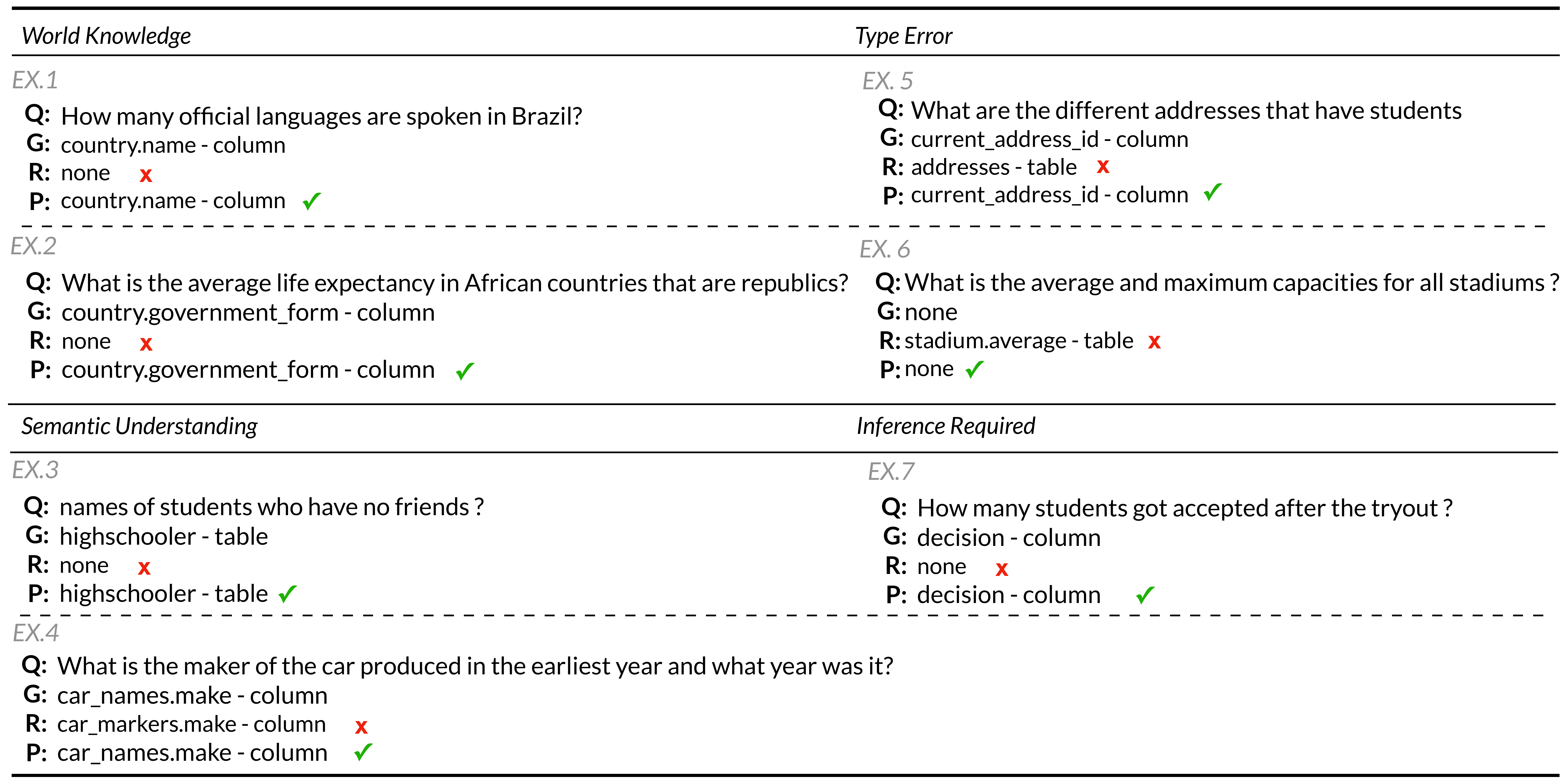}
    \caption{Representative erroneous schema linking predictions by rule-based methods. Notations Q, G, R, and P stand for question, ground truth, rule-based approach prediction and \name prediction, respectively.
    }
    \label{fault_type}
\end{figure*}

\subsection{Schema Linking Performance Analysis}
To have a better analysis on how \name help capture relational knowledge from PLMs, we carefully conduct error checking in terms of schema linking on the Spider benchmark.
We analyze the errors made by previous works \cite{ratsql,lgesql} and classify them into four categories: \textit{World Knowledge Error}, \textit{Semantic Understanding Error}, \textit{Type Error} and \textit{Inference Error}. We observe that \name can successfully solve most of these bad cases which the previous methods fail to address. Due to the limited space, we only report one or two representative examples for each error category  in Figure \ref{fault_type}. From the results, we have the following observations.

First, we find that most of wrong predictions are due to the lack of world knowledge \cite{slsql}. For example, as shown in the first example in Figure \ref{fault_type}, the rule-based semantic linking with exact text matching can not predict ``\textit{Brazil}'' as a country name , and it also fails to understand that ``\textit{republic}'' is a government form in the second example. 
Theoretically, PLMs can be seen as an external knowledge resource by pre-training on large-scale corpora, and previous models with PLMs can identify that the word ``\textit{republic}'' is related to ``\textit{government form}'' by referring to PLMs. However, previous methods with PLMs still fail to capture these reference linking \cite{slsql,ratsql,lgesql}. This is because only using the embeddings of PLMs cannot effectively elicit relational knowledge from PLMs.  
In contrast, our probing method can successfully elicit such world (relational) knowledge, and is portable for practical use.

Second, one kind of error categories is caused by failing to capture semantic relations between words and tables/columns. The rule-based schema linking often chooses columns/tables that exactly occur in the query via exact string matching.
As shown in the fourth example in Figure \ref{fault_type}, the rule-based schema linking method links the word ``\textit{maker}'' to the column ``\textit{car\_makers.Maker}''  while the correct choice should be the column ``\textit{car\_names.Make}''. Similarly, the third example fails to identify that the word ``\textit{highschooler}'' is a synonym of ``\textit{students}''.
This can be solved by considering in-depth semantic information of the query instead of only word occurrence. Our probing method can easily solve this kind of problem.

Third, some questions may contain more than one entity words that can match the table names. The rule-based schema linking could choose the wrong columns that match the entity names occurred in the question. For example, in the fifth example, the rule-based schema linking predicts the wrong column ``\textit{Address}'' instead of the correct column ``\textit{current\_address\_id}''. 
In addition, in some special cases, when the column name exactly matches the keyword in the question,  the rule-based schema linking can not distinguish it clearly, as shown in the sixth example.
This problem often occurs when the question contains keywords such as average, maximum, and minimum, etc.

Finally, the rule-based schema linking cannot solve the difficult cases that require strong inference ability.
Taking the seventh example as an example, the word ``\textit{accepted}'' in the question implicitly links the column ``\textit{decision}'', which cannot be identified by exact string matching. 
Our \name model can solve these difficult cases by using semantic information learned from PLMs.

\begin{table*}[t]
\centering
\small
\setlength{\abovecaptionskip}{5pt} 
\caption{Case study: the first two cases are sampled from SYN and the last two cases are sampled from DK.}
\resizebox{0.75\hsize}{!}{
\begin{tabular}{ll}
\toprule
 Question & \textit{Find the \textbf{\sout{type}} and weight of the youngest pet.} \\
 SYN\_Question& \textit{Find the \textbf{category} and weight of the youngest pet.}\\
 LGESQL & SELECT Pets.\bad{Pet\_age}, Pets.weight FROM Pets ORDER BY Pets.{pet\_age} LIMIT 1  \\
 \name& SELECT Pets.\good{PetType}, Pets.weight FROM Pets ORDER BY Pets.{pet\_age} LIMIT 1 \\
 Gold& SELECT Pets.PetType , Pets.weight FROM Pets ORDER BY Pets.pet\_age LIMIT 1
 \\ \midrule
 Question & \textit{How many distinct \textbf{\sout{countries}} do \textbf{\sout{players}} come from?}\\
 SYN\_Question& \textit{How many distinct \textbf{states} do \textbf{participants} come from?} \\ 
 LGESQL& SELECT COUNT(\bad{*}) FROM \bad{matches}\\
 \name& SELECT COUNT(\good{DISTINCT players.country\_code}) FROM \good{players}\\
 Gold& SELECT COUNT(DISTINCT players.country\_code) FROM {players} \\ 
 \midrule
 Question & \textit{What are the names of the singers who are not French?}\\
 LGESQL & SELECT singer.Name FROM singer WHERE singer.\bad{Name} != 'French' \\
 \name & SELECT singer.Name FROM singer WHERE singer.\good{Citizenship} != 'French' \\
 Gold & SELECT singer.Name FROM singer WHERE singer.Citizenship != 'French'
 \\ \midrule
 Question & \textit{Find the average and maximum id for each type of pet.}\\
 LGESQL & SELECT Pets.PetType, Avg(Pets.\bad{PetType}), Max(Pets.\bad{PetType}) FROM Pets GROUP BY Pets.PetType \\
 \name & SELECT Pets.PetType, Avg(Pets.\good{PetID}), Max(Pets.\good{PetID}) FROM Pets GROUP BY Pets.PetType \\
 Gold & SELECT Pets.PetType, Avg(Pets.PetID), Max(Pets.PetID) FROM Pets GROUP BY Pets.PetType

\\ \bottomrule
\vspace{-0.5cm}
\end{tabular}
}

\label{tab:case}
\end{table*}

\begin{figure*}[h]
    \centering
    \setlength{\abovecaptionskip}{2pt} 
    \subfigure{
             \includegraphics[width=0.23\textwidth]{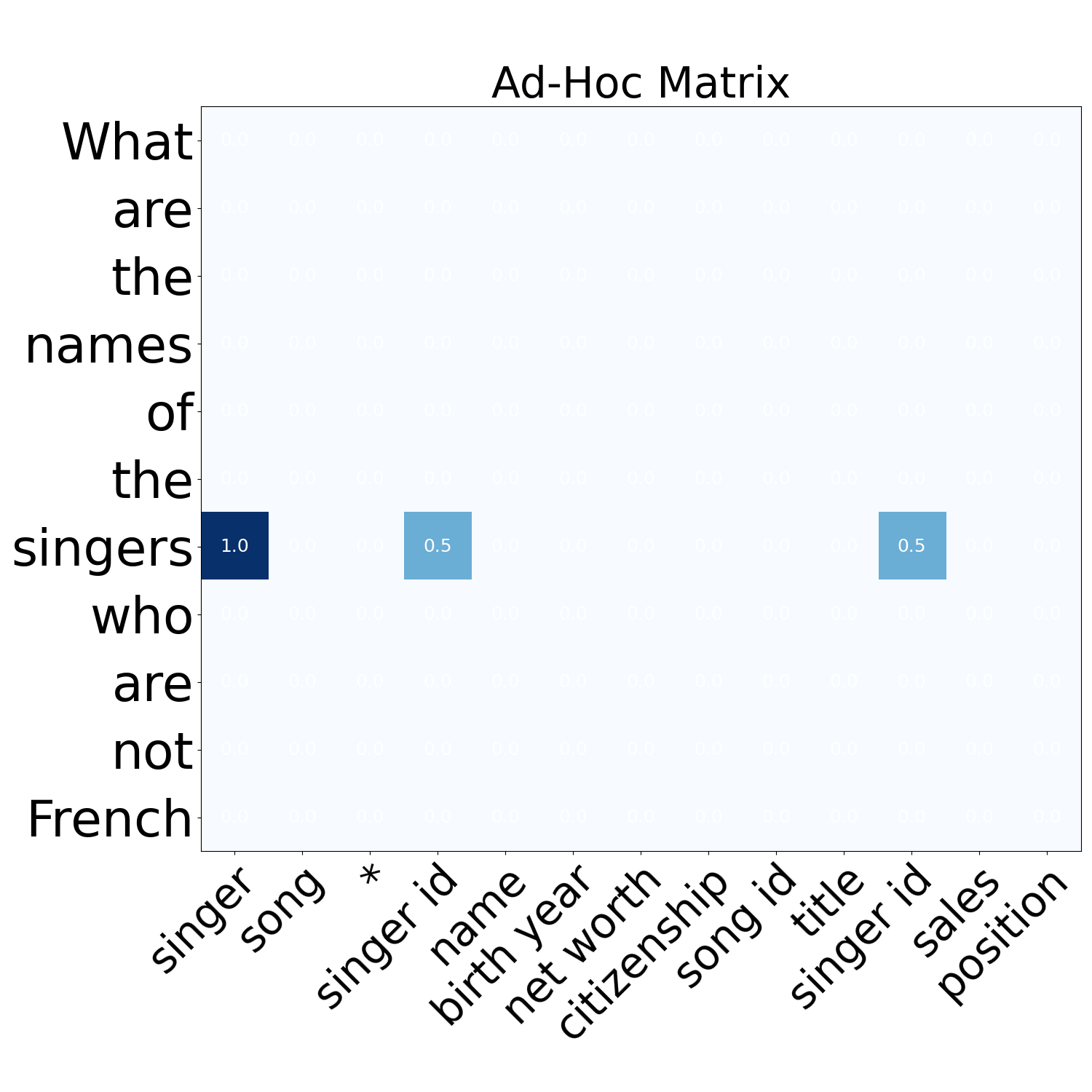}
             \includegraphics[width=0.23\textwidth]{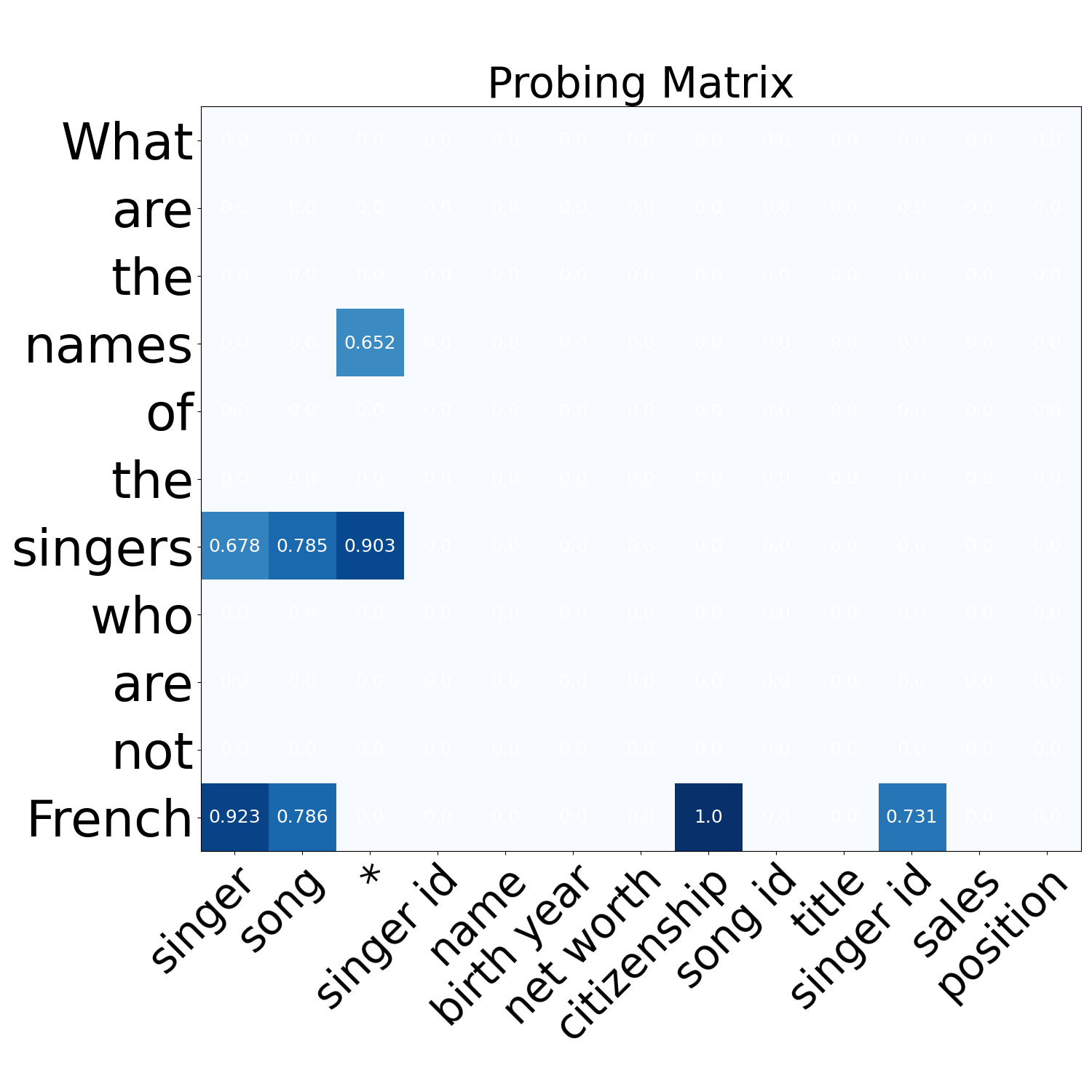}
    }
    \subfigure{
             \includegraphics[width=0.23\textwidth]{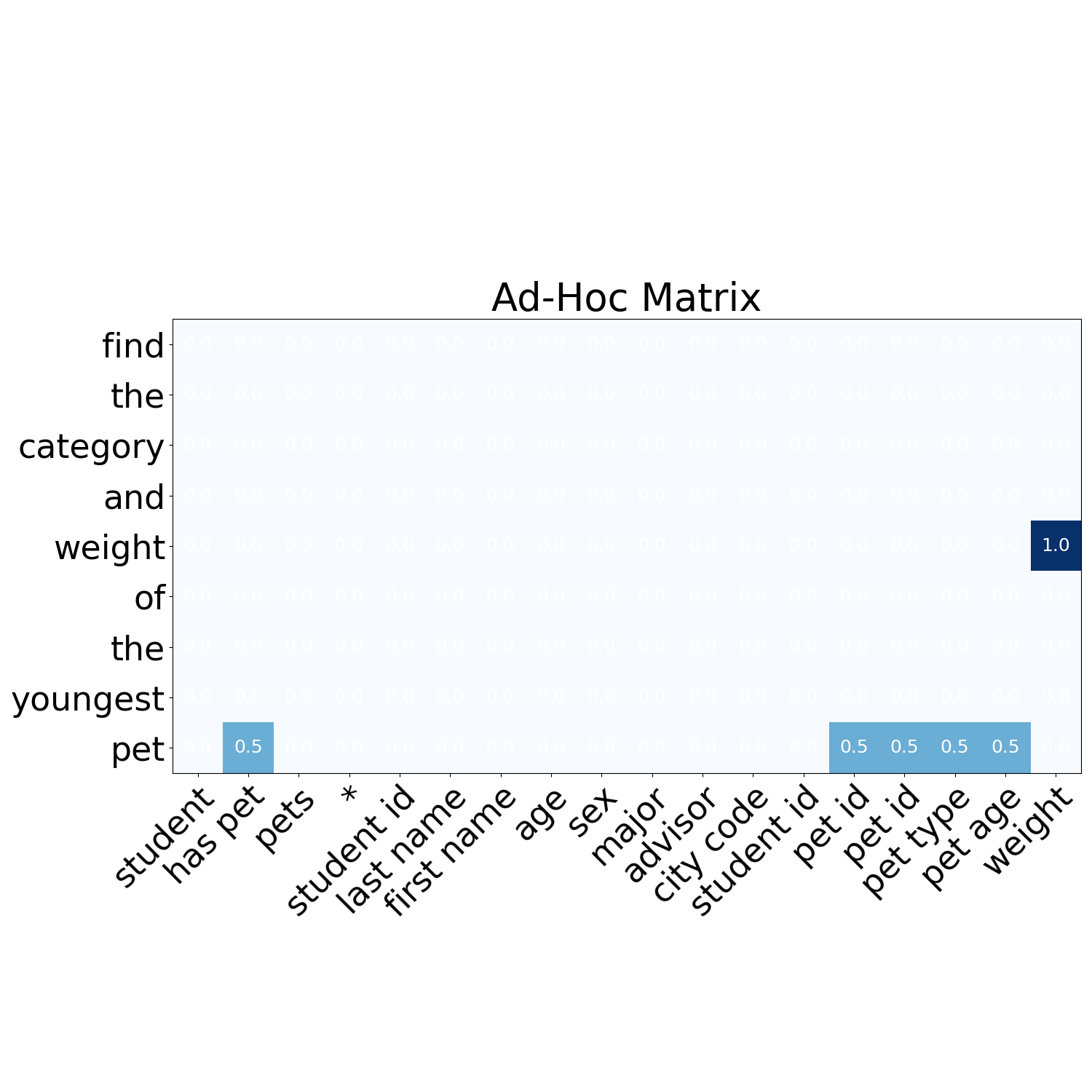}
             \includegraphics[width=0.23\textwidth]{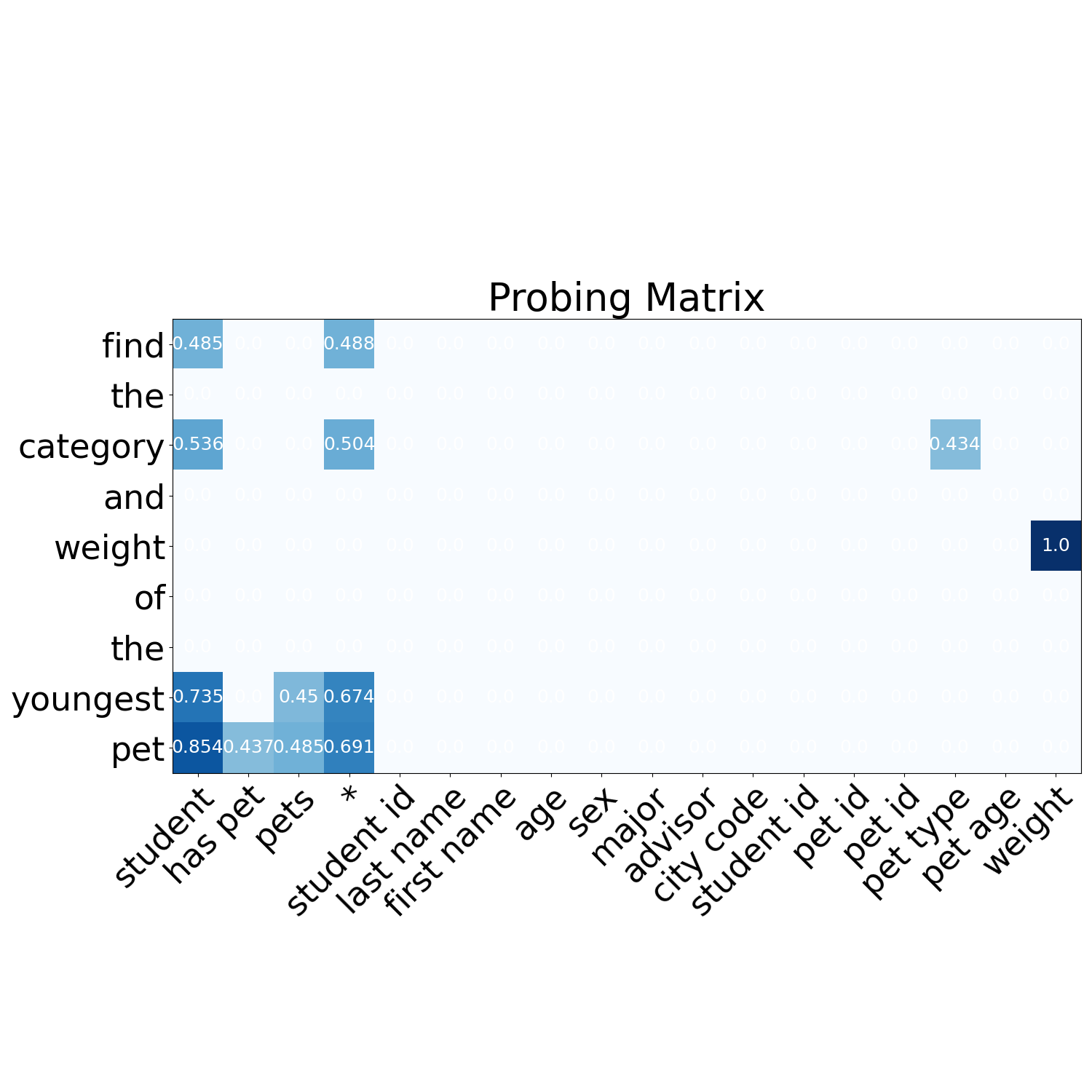}
    }
     \vspace{-0.2cm}
    \caption{The visualization of Rule-based Matrix and Probing Matrix on two cases. The left two sub-figures correspond to the first case (SYN) in the Table \ref{tab:case} and the right two sub-figures correspond to the third case (DK) in the Table \ref{tab:case}.}
    \label{fig:case1}
\end{figure*}

\vspace{-0.2cm}
\subsection{Case Study}
We use four exemplary cases selected from DK and SYN benchmarks to demonstrate the effectiveness of our model qualitatively. Table \ref{tab:case} shows the SQL queries generated by the best baseline LGESQL and our \name model, where the first two cases are from the SYN benchmark and the last two cases are from the DK benchmark. From the results, we can observe that \name can generate correct SQL queries when dealing with challenging scenarios such as synonym substitution.
For instance, in the first case, when replacing the schema-related word  ``\textit{type}'' with its synonym  ``\textit{category}'' in NL question, LGESQL fails to identify the correct column \texttt{PetType}. 

Figure \ref{fig:case1} shows the visualization of correlation matrices obtained by rule-based feature engineering and our probing technique. In the right two sub-figures, the rule-based technique (exact string matching) cannot capture the alignment between the word ``\textit{category}'' and the schema item \texttt{PetType}. While \name can easily catch such semantic similarity with the help of elicited knowledge from PLMs, it thus generates the correct SQL query. 
Similarly, in the left two sub-figures, \name successfully links the domain knowledge word ``\textit{French}'' to column \texttt{Citizenship}, while LGESQL fails to identify the column \texttt{Citizenship} without explicit mentions.
We believe that  \name can probe rich semantic and relational knowledge from large-scale PLMs, facilitating schema linking in text-to-SQL parsing.

\section{Related Work}

\noindent\paragraph{\textbf{Text-to-SQL Parsing.}}
Recently, meaningful advances emerged on the encoder \cite{bogin-representing,chen2021shadowgnn,hui2022ssql,hui2021dynamic}, decoder \cite{YinN17,Choi2020RYANSQLRA,hui2021improving} and table-based pre-training models \cite{Yin2020TaBERTPF,Yu2020GraPPaGP,Shi2020LearningCR,qin2021sdcup,liu2022tapex} on Spider benchmark\cite{spider}.
In particular, \citet{slsql} pointed out that the schema linking module in the encoder was a crucial ingredient for successful prediction. 
To tackle the problem, \citet{typesql} incorporated prior knowledge of column types and schema linking as additional input features. \citet{guo2019towards} used heuristic rules to construct intermediate representation. \citet{editsql} used the co-attention mechanism to measure similarity between NL tokens with schema tokens.
The recent method RAT-SQL \cite{ratsql} utilized a relational graph attention to handle various pre-defined relations and further considered both local and non-local edge features.
To tackle the robustness problem in a more realistic setting, \citet{spiderSYN} proposed to use different data augmentation techniques including data annotation and adversarial training.
\citet{wang2020meta} proposed a model-agnostic meta-learning based training objective to boost out-of-domain generalization of text-to-SQL models.
\citet{Scholak2021PICARDPI} propose PICARD, a method for constraining auto-regressive decoders of language models through incremental parsing. 
Different from these methods, we are the first to explore the potential knowledge stored in PLMs to help the model perform better schema linking and further improve the generalization of the model.

\noindent\paragraph{\textbf{Probing PLMs.}}
The success of PLMs has led to a large number of studies investigating and interpreting the rich knowledge that PLMs learn implicitly during pre-training \citep{kovaleva-etal-2019-revealing, rogers2020primer,he2021galaxy, he2022unified}.
One typical approach is to probe PLMs with a small amount of learnable parameters considering a variety of linguistic properties, including morphology \citep{belinkov-etal-2017-neural}, word sense \citep{bert-geomertry}, syntax \cite{hewitt2019structural, dai2021does}, world knowledge \cite{petroni2019language} and semantics \citep{liu-etal-2019-linguistic}.
Another line of work is motivated to probe PLMs in an unsupervised and parameter-free fashion \citep{wu2020perturbed, li-etal-2020-heads}.
Our work generally follows this line and exploits an unsupervised probing technique to extract relational knowledge for the downstream text-to-SQL parsing task.
\citet{liu2021awakening} proposed the ETA model to explore the grounding capabilities of PLMs. The proposed erasing-then-awakening  trains a concept classification module by human-crafted supervision.
Then, it erases tokens in a question to obtain the concept prediction confidence differences as pseudo alignment.
Finally, it awaken latent grounding from PLMs by applying pseudo alignment as supervision.
This method requires human-crafted label as supervision, which could not be easily obtained in most tasks.
Furthermore, the additional trainable parameters may cause failures to adequately reflect differences in representations \citep{hewitt-liang-2019-designing}.

Different from previous methods, \name does not need any extra labels or supervision, which is not limited to specific tasks.
Our method follows an unsupervised technique, which makes sure all the relational knowledge is extracted from PLMs.
In addition, \name utilizes a direct graph to represent the relational knowledge and can better extract the relational information within the input.

\section{Conclusion}
In this paper, we proposed a probing technique to probe schema linking information between the NL query and the database schema from large-scale PLMs, which improved the generalization and robustness of the text-to-SQL parsing models. In addition, a Poincar\'e distance metric was devised to measure the difference between two vectors in the hyperbolic space, capturing the heterogenous relational structures between the NL query and the database schema.
Experimental results on three benchmark datasets demonstrated that our method substantially outperformed strong baselines and set state-of-the-art performance on three text-to-SQL benchmarks.

\begin{acks}
This work was partially supported by National Natural Science Foundation of China (No. 61906185), Youth Innovation Promotion Association of CAS China (No. 2020357), Shenzhen Science and Technology Innovation Program (Grant No. KQTD20190929172835662), Shenzhen Basic Research Foundation (No. JCYJ20210324115614039 and No. JCYJ20200109113441941).
This work was supported by Alibaba Group through Alibaba Innovative Research Program.
\end{acks}

\bibliographystyle{ACM-Reference-Format}
\bibliography{anthology.bib,custom.bib,text2sql.bib}

\appendix

\begin{table*}[]
\centering
\small
\setlength{\abovecaptionskip}{5pt} 
\caption{Exact matching accuracy by varying the levels of difficulty of the inference data on the development sets of DK, SYN and Spider.}
\resizebox{1.0\hsize}{!}{
\begin{tabular}{l|ccccc|ccccc|ccccc}
\toprule
\multirow{2}{*}{Model} & \multicolumn{5}{c|}{DK}             & \multicolumn{5}{c|}{SYN}            & \multicolumn{5}{c}{Spider}         \\
\cline{2-16}
        & easy & medium & hard & extra & all & easy & medium & hard & extra & all & easy & medium & hard & extra & all \\
\midrule
RAT-SQL     &69.0 &42.2 &18.9 & 11.4& 38.5   &68.9 &57.5 &32.2 &15.9 & 49.1  & 87.9 & 74.6 &60.3 & 48.7 & 71.2 \\
\midrule
RAT-SQL+Euclidean \name  &69.0 &45.5 &31.0 &\textbf{29.5} & 45.2   & \textbf{80.2}&62.9 &51.4 &\textbf{40.2} &61.4     & 86.2 &74.6  &64.9 &\textbf{54.8} & 72.6   \\
\midrule
RAT-SQL+Hyperbolic \name &\textbf{71.8}  &\textbf{46.3} &\textbf{33.8} &28.6 & \textbf{46.4}    & 78.2    &\textbf{66.8}    & \textbf{54.2}     &  37.9     & \textbf{62.6}  & \textbf{88.3}     & \textbf{76.0}   & \textbf{65.5}    & 50.0     &  \textbf{73.1}   \\
\midrule
\midrule
LGESQL     & 74.5     & 46.7      & \textbf{41.9}    & 29.5      &  48.4   &  79.4    &  \textbf{67.9}      & 62.1     & 36.1      & 64.6    &  91.9    &  78.3      &  64.9    &  \textbf{52.4}    & 75.1    \\
\midrule
LGESQL+Euclidean \name & 72.8     & 49.6       & 40.5     & 31.4      & 49.3    & 81.5     & 67.3       & \textbf{62.1}     & 39.8      & 65.4    & 91.9     & 79.4       & 70.1     & 50.0      & 76.0    \\
\midrule
LGESQL+Hyperbolic \name & \textbf{75.5}     & \textbf{50.8}       & 40.5     &  \textbf{33.3}     &  \textbf{51.0}   &  \textbf{81.9}    &  67.3      & 60.9     &   \textbf{41.6}    &  \textbf{65.6}   & \textbf{92.7}     &  \textbf{79.6}      & \textbf{68.4}     &51.2       &  \textbf{76.3}  \\
\bottomrule

\end{tabular}
}
\label{diff_result}
\end{table*}

\section{Results on Complex Queries}
The three benchmarks provide four different difficulty levels of samples. We investigate the detailed model performance and have further insights on how \name can help complex queries.
Table \ref{diff_result} shows the exact match accuracy by varying the levels of difficulty of the data.
From the results, we can observe that  \name can boost the performance of SOTA text-to-SQL parsers (RAT-SQL/LGESQL) across almost all different difficulty levels on the three benchmark datasets.
It suggests that \name can lead to more significant accuracy improvements compared to RAT-SQL/LGESQL.
For example, LGESQL with Hyperbolic \name, which better captures the relational knowledge, achieves the highest score on most cases.
In particular, \name gains much better performance on the extremely hard samples than the baselines, verifying that the harder samples require better schema linking for correct text-to-SQL parsing.

\begin{table}
\centering
\small
\setlength{\abovecaptionskip}{5pt} 
\caption{Comparison of the inference time in seconds on 1034 SYN samples.}
\scalebox{1}{
\begin{tabular}{lccc}
\toprule
  & Model & \multicolumn{1}{l}{Model+Euclidean} & Model+Hyperbolic        \\
\midrule
LGESQL &  878(s)   & 979(s)                                 & \multicolumn{1}{c}{975(s)} \\
RAT-SQL &   1162(s)       & 1255(s)                                &\multicolumn{1}{c}{1387(s)}\\
\bottomrule
\end{tabular}
}
\label{tab:time}
\end{table}

\section{Computational Cost}
We investigate the computational cost of baseline methods and our \name model in inference.
All these models are run on a desktop machine with a single NVIDIA Tesla V100 GPU. 
In Table \ref{tab:time}, we report the inference time on 1034 SYN samples with the batch size of 1. \name has a slightly slower inference speed than the base models (LGESQL and RAT-SQL). For example, on average, the inference time of \name with Poincar\'e probe increases by 0.09s on each sample compared with LGESQL, which is acceptable in practice.

\end{document}